\documentclass[runningheads]{llncs}
\usepackage[T1]{fontenc}
\usepackage{graphicx}
\usepackage[linesnumbered,ruled,vlined]{algorithm2e}
\usepackage{multirow}
\usepackage{comment}
\usepackage{float}
\usepackage{color}
\usepackage{booktabs}
\usepackage{amsmath}
\begin{document}
\title{Mining Large Independent Sets on Massive Graphs}

\author{
Yu Zhang\inst{1} \and
Witold Pedrycz\inst{2} \and
Chanjuan Liu\inst{3} \and
Enqiang Zhu\inst{4}
}
\authorrunning{Y. Zhang et al.}

\institute{
Peking University, Beijing, China\\
\email{yuzhang.cs@stu.pku.edu.cn}\\
\and
University of Alberta, Canada\\
\email{wpedrycz@ualberta.ca}\\
\and
Dalian University of Technology, China\\
\email{chanjuanliu@dlut.edu.cn}
\and 
Guangzhou University, China\\
\email{zhuenqiang@gzhu.edu.cn}
}

\maketitle
\begin{abstract}
The Maximum Independent Set problem is fundamental for extracting conflict-free structure from large graphs, with applications in scheduling, recommendation, and network analysis. However, existing heuristics can stagnate when search schedules are fixed and information from past solutions is underused, leading to wasted effort in low-quality regions of the search space. 
We present ARCIS, an efficient algorithm for mining large independent sets on massive graphs. ARCIS couples two main components. The first is an adaptive restart policy that refreshes exploration when progress slows. The second is Consensus-Guided Vertex Fixing, which restricts the search to the non-consensus region of the graph by fixing vertices consistently observed within a round. The consensus is maintained as a running intersection within each round, and because it is recomputed at every restart, the fixing is reversible. Vertices that later lose support are automatically unfixed and their neighborhoods re-enter the working graph, which corrects occasional mistakes while preserving progress.
Experiments on 222 graphs from four benchmark suites show that ARCIS attains the best or tied-best solution quality in most instances while delivering competitive runtime and low variability. Ablation studies isolate the impact of each component, indicating that ARCIS is a practical and robust method for large-scale graph mining.
\end{abstract}
\section{Introduction} \label{sec:intro}
In an undirected graph \( G = (V, E) \), an \emph{independent set} (IS) \( I \subseteq V \) is defined as a subset of vertices where no two vertices are adjacent. The \emph{maximum independent set} (MIS) problem aims to determine an independent set with the largest cardinality, which is one of the first graph theory problems proven to be NP-complete~\cite{karp1972reducibility}. 
The MIS problem is crucial in graph theory due to its computational complexity and is a foundational task in data mining and graph analytics, with extensive applications in fields like wireless network analysis~\cite{alzoubi2003maximal,bai2015maximal}, scheduling~\cite{joo2015distributed,eddy2021maximum}, and bioinformatics~\cite{aviram2012efficient,bulteau2021new}. For example, a common scheduling scenario can be represented by constructing a graph in which the vertices signify resources and the edges represent conflicts between those resources. Identifying the largest subset of resources that can coexist without conflicts is equivalent to solving the MIS problem.

The MIS problem has been extensively studied, leading to both exact exponential algorithms and efficient inexact algorithms. While polynomial-time algorithms are available for bipartite graphs, no such algorithms exist for general graphs. Exact MIS algorithms aim to improve the upper bounds on exponential running times. Currently, the most advanced exact algorithms utilize the measure-and-conquer strategy, with the fastest polynomial-space algorithm achieving a running time of \(1.1996^n n^{O(1)}\)~\cite{exact1}. However, exact exponential-time algorithms can prove impractical for large instances due to their significant computational costs.

Heuristic approaches are widely utilized to obtain high-quality independent sets in a reasonable time. 
One of the particularly successful heuristic algorithms for addressing the MIS problem is ARW~\cite{andrade2012fast}. ARW is founded on the Iterated Local Search (ILS) metaheuristic~\cite{lourencco2003iterated}, with each iteration comprising two phases: a perturbation phase and a local search phase. In this context, ARW employs a perturbation strategy that randomly adds \( k \) (typically \( k=1 \)) non-solution vertices to the current independent set, subsequently removing their neighboring vertices from the solution. This approach enables ARW to avoid local optima and effectively navigate various regions of the solution space. 

Building upon ARW, many advanced MIS algorithms have emerged. For example, Lamm et al. integrated ARW with evolutionary algorithms and a series of exact reduction rules, as implemented in VCSolver~\cite{akiba2016branch}, to introduce ReduMIS~\cite{lamm2016finding}. This algorithm is specifically designed to identify independent sets in large-scale sparse instances. While ReduMIS demonstrates high accuracy, it is constrained by considerable preprocessing overhead. To tackle this limitation, Dahlum et al. introduced two accelerated local search algorithms: KerMIS and OnlineMIS~\cite{dahlum2016accelerating}. KerMIS begins by employing VCSolver to reduce the graph, followed by an inexact reduction rule based on the premise that the probability of an MIS containing high-degree vertices is minimal. It then applies the ARW algorithm to process the remaining subgraph. Conversely, OnlineMIS begins by simplifying the graph through straightforward reduction rules, creating a reduced graph. It then applies an inexact reduction technique and utilizes a variant of the ARW algorithm on the simplified subgraph. These improvements lead to more efficient and scalable solutions, especially for large and complex MIS instances. Chang et al. proposed a reducing-peeling framework for the MIS problem, utilizing exact reduction rules on low-degree vertices and inexact reduction rules on high-degree vertices (when exact reductions are not feasible) to iteratively reduce the graph~\cite{chang2017computing}. Building on this framework, they presented a series of algorithms for MIS, including two baseline algorithms---BDOne and BDTwo---alongside a linear algorithm called LinearTime, a near-linear algorithm NearLinear, and ARW-accelerated algorithms, such as ARW-NL. Hespe et al. developed an advanced parallel kernelization algorithm for the MIS, known as ParFastKer (which utilizes 32 threads) and FastKer (which operates with a single thread). This algorithm integrates parallelization techniques with dependency checking and reduction tracking to enhance the effectiveness of ARW~\cite{hespe2019scalable}. 

In recent years, Gu et al. proposed a reduction-based heuristic algorithm, HtWIS, for solving the maximum weighted independent set (MWIS) problem~\cite{gu2021towards}. The method begins by simplifying the problem instance through exhaustive reductions and applying a tie-breaking policy to select a vertex for inclusion in the solution. Then, the selected vertex and its neighbors are removed from the graph, and the reduction process is repeated. Gro{\ss}mann et al. introduced an advanced memetic algorithm, M$^2$WIS, for the MWIS problem. This approach combines data reduction techniques with a memetic strategy, recursively selecting vertices likely to belong to a high-weight independent set. They also developed a variant, M$^2$WIS+S, incorporating the initial solution construction algorithm CyclicFast~\cite{gellner2021boosting} to generate individuals for the initial population, enhancing the algorithm's performance~\cite{grossmann2023finding}. In addition, CHILS adopts a concurrent hybrid iterated local search that maintains multiple solutions in parallel. At checkpoints it builds a \emph{Difference Core} from solution disagreements and alternates improvement between this core and the full graph, and the implementation supports both MWIS and MIS~\cite{grossmann2025concurrent}.

\paragraph{Motivation.}
Despite substantial progress, MIS heuristics on large graphs can stall. Fixed restart schedules and one-shot reductions may leave the search in low-quality regions, and signals present in recent solutions are rarely reused. Our goal is twofold: refresh exploration when progress slows, and turn persistent agreement across recent solutions into a \emph{reversible} mechanism that \emph{fixes} vertices and thereby constrains subsequent search to the \emph{non-consensus} region.

\paragraph{Contributions.}
We present ARCIS (Adaptive Restarts with Consensus-Guided Vertex Fixing for Independent Set), an efficient algorithm that couples an adaptive restart policy with Consensus-Guided Vertex Fixing. Unlike difference-core and parallel kernelization pipelines that construct (near-)static cores and irreversibly prune regions before search, ARCIS maintains a running intersection within each round and commits only a capped subset at round boundaries with explicit revocation; it can operate on top of any reduced kernel.
\begin{enumerate}
\item[(i)] \emph{Adaptive restart policy.} ARCIS adapts restarts to observed progress rather than following a preset schedule, limiting time spent in low-quality regions and improving the exploration–exploitation balance.
\item[(ii)] \emph{Consensus-Guided Vertex Fixing (CGVF).} Within each round, ARCIS maintains a running intersection of high-quality solutions; at the next restart boundary, vertices supported by this consensus are \emph{fixed}, and the working graph for the following round is rebuilt accordingly. Subsequent search is restricted to the non-consensus region, and vertices that later lose support are automatically reintroduced, yielding a reversible and error-tolerant mechanism.
\item[(iii)] \emph{Empirical validation.} On 222 graphs from four real-world benchmarks, ARCIS attains the best or tied-best solution quality on most instances while delivering competitive runtime and low variability across seeds. Ablation studies isolate the impact of each component, indicating that ARCIS is a practical and robust method for large-scale graph mining.
\end{enumerate}

\section{Preliminaries} \label{sec:preliminaries}

This section introduces some necessary notations and terminologies. Additionally, since the algorithm proposed in this paper integrates the ARW algorithm, a brief review of ARW~\cite{andrade2012fast} is provided.

\subsection{Notations and Terminologies}
This paper considers only unweighted, undirected, and simple graphs. Given a graph \( G \), we denote its \emph{vertex set} and \emph{edge set} as \( V(G) \) and \( E(G) \), respectively. For any vertices \( u, v \in V(G) \), if \( uv \in E(G) \), then \( u \) and \( v \) are referred to as \emph{adjacent} in \( G \), with one vertex being a \emph{neighbor} of the other.
The set of all neighbors of a vertex \( u \in V(G) \) is known as the \emph{neighborhood} of \( u\), denoted by \( N_G(u) \). The \emph{closed neighborhood} of \( u\), denoted by \( N_G[u] \), is defined as \( N_G[u] = N_G(u) \cup \{u\} \). The cardinality of \( N_G(u) \) is called the \emph{degree} of \( u \) in \( G\), denoted by \( d_G(u) \). A vertex with degree \( k \) is called a \emph{\( k \)-vertex}. For a subset \( S \subseteq V(G) \), we define: \( N_G(S) = \bigcup_{v \in S} N_G(v) \setminus S \) and \( N_G[S] \)= \( N_G(S) \cup S \). To \emph{delete} a set \( S \) from \( G \) means removing all vertices in \( S \) along with their incident edges, resulting in the graph denoted by \( G - S \). When \( S \) consists of a single vertex \( \{u\} \), it is simplified to \( G - u \) for convenience. Furthermore, \( G - S \) is referred to as the \emph{subgraph induced} by \( V(G) \setminus S \), which can also be expressed as \( G[V(G) \setminus S] \).

An \emph{independent set} of a graph \( G \) is a subset of vertices \( I \subseteq V(G) \) such that no two vertices in \( I \) are adjacent in \( G \). An independent set \( I \) is termed 
 \emph{maximum} if there is no independent set \( I' \) such that $|I|< |I'|$. The Maximum Independent Set (MIS) problem seeks to identify an MIS within a given graph. The size of an MIS in \( G \) is referred to as the \emph{independent number} of $G$, denoted by \( \alpha(G) \). While a graph \( G \) may contain multiple MISs, accurately determining \( \alpha(G) \), whether exactly or approximately, is a challenging endeavor. This complexity underlines the importance of our research into the MIS problem. We aim to design an efficient heuristic algorithm to find a high-quality independent set of a graph.

\subsection{The Local Search algorithm ARW}

ARW extends the concept of swap to a \((j, k)\)-swap, which involves removing \(j\) vertices from the current solution and adding \(k\) vertices.
One iteration of ARW consists of one perturbation and one local search step. In the perturbation step, ARW randomly selects vertices that have been out of the solution for the longest time, adds the vertices to the current solution, and removes their neighbors from the current solution. In most cases, only one vertex is forced into the current solution. However, in infrequent scenarios, multiple vertices may also be incorporated into the solution, where the number of forced vertices is $i + 1$ with probability $1/2^i$. In $O(|E|)$-time, ARW can find a (1,2)-swap or prove that a (1,2)-swap does not exist, where $E$ is the edge set of the input graph. ARW uses (1,2)-swaps in the local search phase to improve the current solution.

\section{The ARCIS Algorithm} \label{sec:arcs}

ARCIS is an iterative local search algorithm that operates in rounds and maintains two key state sets with distinct lifecycles: a committed consensus set $S$ and a running intersection set $H$. The committed consensus $S$ is fixed throughout each round and defines the working graph on which local search operates, while the running intersection $H$ is updated periodically within the round. When a restart triggers a new round, the accumulated evidence in $H$ is committed to $S$ for the next round. In this way, each round operates with a consensus $S$ fixed at its start, and the next round inherits the evidence gathered during the current round. Algorithm~\ref{alg:arcs} outlines the overall procedure of ARCIS.

\begin{algorithm}[htbp]
\KwIn{Graph $G=(V,E)$, the \emph{cutoff} time, checkpoint interval $c$}
\KwOut{An independent set $I^*$ of $G$}
$I^*:=\emptyset$, $S:=\emptyset$, $H:=\emptyset$, $restart:=true$, $Step:=0$, $p:=0$\;
$(G,I_s^*):= $\textsc{Preprocessing}$(G)$\;
\While{elapsed\_time $<$ cutoff}{
  \If{$restart$}{
    $S:=H$, $H:=\emptyset$, $Step:=0$, $restart:=false$\;
    $K:=G-N[S]$\;
    $I:=$\textsc{ConstructIS}$(K)$\;
  }
  $I:=ARW(K,I)$\;
  $Step:=Step+1$\;
  \If{$Step\bmod c=0$}{
    \If{$H=\emptyset$}{$H:=I$}\Else{$H:=H\cap I$}
    \If{$|I_s^*|+|S|+|I|>\,|I^*|$}{$I^*:=I_s^*\cup S\cup I$}
    \Else{$(restart,p):=$\textsc{Evaluation}$(Step,p)$\;}
  }
}
\Return{$I^*$}\;
\caption{\textsc{ARCIS}}
\label{alg:arcs}
\end{algorithm}

ARCIS begins by initializing all states and performing a one-time Preprocessing step~\cite{akiba2016branch} that reduces the input graph, producing a smaller graph $G$ and a partial independent set $I_s^*$ from the removed vertices (lines 1--2). This preprocessing yields a fixed initial solution component and significantly reduces the search space. In practice, we adopt the exact kernelization as implemented in ReduMIS~\cite{lamm2016finding}.

Each subsequent round of ARCIS starts by committing the previous round’s intersection $H$ into the consensus set $S$, then clearing $H$ for fresh accumulation (line 5). We reset the local step counter and the restart flag, and rebuild the working graph $K$ by removing $S$ and its neighborhood from the original graph $G$ (line 6). 
Because $K$ is always rebuilt from the original reduced graph $G$ as $K := G \setminus N[S]$, vertices revoked by $S:= H$ naturally have their neighborhoods reintroduced into $K$ at the start of the next round, which realizes the intended reversibility.
Next, ARCIS constructs an initial independent set $I$ on the working graph $K$ via the routine \textsc{ConstructIS} (line 7). This initializer uses a simple greedy strategy: it repeatedly selects a vertex of minimum degree from $K$, adds it to $I$, and removes that vertex’s closed neighborhood from $K$. This process continues until no vertices remain in $K$. Because this construction operates only on the reduced graph $K$, the resulting set $I$ is a valid independent set in $K$ (hence also feasible in the original graph once combined with $S$ and $I_s^*$). At this point, the algorithm hands $I$ over to the ARW local search procedure for further improvement.

After initialization, ARCIS runs ARW on the working graph $K$ (line~8). Each call at line~8 performs exactly one ARW local iteration, and we denote by \emph{Step} the total number of ARW iterations since the last restart.
At each checkpoint (every $c$ iterations; lines~10--18), it updates the running intersection $H$ by intersecting the current solution $I$ with the round accumulator ($H:=I$ at the first checkpoint, otherwise $H:=H\cap I$). In the same block, ARCIS evaluates the best solution on the \emph{original} graph by combining the preprocessing part $I_s^*$, the committed consensus $S$, and the current local solution $I$; if $|I_s^*|+|S|+|I|$ improves over $|I^*|$, it updates $I^*$.
If no improvement occurs at a checkpoint, \textsc{Evaluation} may trigger a restart (line~18). A restart ends the round, commits the accumulated evidence ($S\leftarrow H$), and rebuilds the next working graph $K:=G-N[S]$; otherwise the search continues on the current $K$ with the same $I$. The process repeats until the cutoff is reached, at which point ARCIS returns the best-found independent set $I^*$ (line~19).

\paragraph{Main-loop complexity.}
In each round, we rebuild the working graph $K$ and construct an initial set $I$ on $K$ via \textsc{ConstructIS}, which takes $O(|V(K)| + |E(K)|)$ time per round.
The local search then performs ARW \emph{iterations} on $K$; the per-iteration cost is worst-case $O(|E(K)|)$~\cite{andrade2012fast}.
Every $c$ iterations, a checkpoint updates the running consensus $H$ by one set intersection, which is $O(|V(K)|)$ under standard set representations, and \textsc{Evaluation} is $O(1)$.
Under a fixed wall-clock cutoff, the runtime is therefore dominated by the number and per-iteration cost of ARW iterations executed before the cutoff, while the checkpoint and evaluation add only small periodic overhead proportional to the number of checkpoints.

\section{Main Components of ARCIS} \label{sec:components}

The effectiveness of ARCIS stems from two core innovations that guide the local search process: an adaptive restart policy that dynamically determines when to restart the search, and a Consensus-Guided Vertex Fixing (CGVF) strategy that progressively fixes certain vertices based on consensus. In this section, we describe these two components in detail and explain how each contributes to ARCIS’s performance.

\subsection{Adaptive restart policy} \label{sec:adaptive}
Rather than using a fixed periodic restart schedule, ARCIS employs an adaptive restart policy that responds to observed progress. The decision is evaluated only at checkpoints in Algorithm~\ref{alg:arcs} (every $c$ iterations) and only when no improvement is found at that checkpoint, using the procedure in Algorithm~\ref{alg:adaptive_restart}. The goal is to keep restarts rare while the search is productive and to increase their likelihood as stagnation persists.

The procedure maintains a restart probability $p$. When the internal probe condition $Step\bmod n = 0$ holds (line~1), it triggers a restart with probability $p$ (lines~2--3); otherwise no restart occurs and $p$ is increased by a fixed increment $\alpha$ (lines~4--5), making a future restart more likely. 
After a restart, $p$ is reset to $0$, allowing the new round to exploit before being probed again. Because \textsc{Evaluation} is bypassed whenever a new best solution is found, restarts remain infrequent while the search is making progress; under persistent stagnation, $p$ accumulates over probes and a restart is eventually triggered. The overhead is minimal, consisting of a constant-time check at conditional checkpoints.

\begin{algorithm}[t]
    \SetKw{Continue}{continue}
    \KwIn{Search iterations $Step$, restart probability $p$}
    \KwOut{Restart flag $restart$ and restart probability $p$}

    \If{$Step \bmod n = 0$}{
        \If{$rand(0,1) < p$}{
            $restart := true$; $p := 0$\;
        }
        \Else{
            $restart := false$; $p := p + \alpha$\;
        }
    }
    \Else{
        $restart := false$\;
    }

    \Return{$(restart, p)$}

    \caption{Evaluation($Step, p$)}
    \label{alg:adaptive_restart}
\end{algorithm}

\subsection{Consensus-Guided Vertex Fixing} \label{sec:inference}
Consensus-Guided Vertex Fixing (CGVF) fixes vertices that persist across solutions within a round so that the next round searches only the remaining subgraph. In Algorithm~\ref{alg:arcs}, the running intersection is updated at checkpoints by intersecting the current solution with the round accumulator (lines~10--14). When a restart occurs, the accumulated evidence is committed as the round consensus, and the working graph for the next round is rebuilt by removing the committed vertices and their closed neighborhoods (lines~4--7).

The mechanism is reversible and conservative. The consensus is recomputed every round, and a previously fixed vertex that no longer appears in the running intersection is not committed again, so its neighborhood returns to the working graph. Solution quality is evaluated on the original graph at checkpoints by combining the preprocessing contribution, the committed consensus, and the current local solution (lines~15--16). The per-checkpoint cost is a single set intersection that is negligible compared with local search, while the adaptive restart policy determines when commitment takes effect.

\section{Experimental Evaluation}\label{sec:experiments}


\subsection{Experiment Setup} \label{setup}

We compare ARCIS\footnote{The source code of ARCIS will be made available after the paper is accepted.} against four state-of-the-art heuristics, namely ReduMIS~\cite{lamm2016finding}\footnote{https://github.com/KarlsruheMIS/KaMIS}, HtWIS~\cite{gu2021towards}\footnote{https://github.com/mwis-abc/mwis-source-code}, M$^2$WIS+S~\cite{grossmann2023finding}\footnote{https://github.com/KarlsruheMIS/KaMIS/tree/master/mmwis}, and CHILS~\cite{grossmann2025concurrent}\footnote{https://github.com/KennethLangedal/CHILS}. 
We intentionally focus on these recent strong baselines together with \emph{ReduMIS}, a widely used ARW-based reference in the MIS literature. We include ReduMIS as the representative ARW-family method because it is the most competitive and most frequently used comparator on massive graphs, and its kernelization+ARW design subsumes the plain ARW backbone.
To run methods that also target MWIS on the unweighted MIS task, we assign unit weights to all vertices~\cite{jiang2023exact} and adopt the authors’ recommended unweighted settings. 
All algorithms are executed \emph{single-threaded} with a uniform per-instance wall-clock cutoff of 1000\,s; timing starts after graph loading and includes preprocessing and all subsequent computation.
HtWIS does not rely on random seeds and is run once per instance, whereas all other methods are run ten times with seeds 1--10.

ARCIS exposes two parameters in \textsc{Evaluation}, the probe interval $n$ and the probability increment $\alpha$.
We tune them once with \emph{irace}~\cite{irace} on a training set formed by a randomly sampled 10\% subset from each benchmark (disjoint from the test set), using a budget of 250 runs with a 1000\,s cutoff per run.
The search space is $n \in \{5,10,15,20,25\}\!\times\!10^4$ and $\alpha \in (0,0.1)$; the selected values are $n=2.0\!\times\!10^5$ and $\alpha=0.004$, which are fixed for all experiments.
The checkpoint interval is set to $c=10{,}000$, following ARW-family practice to avoid checking every iteration~\cite{andrade2012fast,lamm2016finding,hespe2019scalable}.

Experiments are conducted on Ubuntu~20.04.1 with an Intel Xeon W7-3445 CPU at 3.40\,GHz and 128\,GB RAM.
All implementations are in C++ and compiled with g++~9.4.0 using `-O3'.

\paragraph{Datasets.}
We evaluated all methods on four well-known benchmarks, commonly used to assess heuristic algorithms for various problems. All instances are considered unweighted, undirected simple graphs. 

\begin{itemize}
    \item Walshaw~\cite{soper2004combined}: This benchmark was originally developed as part of a project focused on studying high-quality graph partitions\footnote{https://chriswalshaw.co.uk/partition/}. It consists of sparse graphs with vertex counts ranging from \(10^3\) to \(10^5\), and has been employed to test algorithms such as ReduMIS \cite{lamm2016finding}.

    \item SNAP~\cite{snapnets}: This benchmark is part of the Stanford Large Network Dataset Collection\footnote{http://snap.stanford.edu/data}. The instances selected from this dataset for our experiments include graphs with vertex counts varying from \(10^3\) to \(10^6\).

    \item DIMACS10~\cite{bader201110th}: This benchmark includes a diverse range of synthetic and real-world graphs from the 10th DIMACS Implementation Challenge, which focuses on graph partitioning and clustering\footnote{https://www.cc.gatech.edu/dimacs10/downloads.shtml}. These graphs are generally sparse, with vertex counts ranging from \(10^4\) to \(10^6\).

    \item Network Repository~\cite{rossi2015network}: This benchmark features graphs derived from various real-world applications\footnote{https://networkrepository.com}. The instances span multiple categories, with vertex counts ranging from \(10^2\) to \(10^7\).
\end{itemize}

\subsection{Comparison of Solution Quality}

We evaluate solution quality using the independent-set size. Per the protocol in Section~\ref{setup}, we report for each algorithm and instance two metrics: `Max', the largest size observed over the runs under the common budget, and `Avg', the arithmetic mean over those runs. For consistency, each table entry is shown as `Max (Avg)'; when `Max = Avg' for an algorithm, we collapse the cell to a single value. The per-instance results are listed in Tables~\ref{tab:Walshaw}--\ref{tab:real-world}; in these tables, the best `Max' and the best `Avg' per instance are shown in bold, which includes tied-best cases. Instances in which all algorithms achieve identical `Max' and `Avg' are omitted to save space.

Across the 222 instances, ARCIS attains the best or tied-best `Max' on 213 instances, with 31 being strictly best (strictly greater than all other algorithms on the same instance). The corresponding counts for the baselines are ReduMIS 185 best `Max' with 5 strictly best, CHILS 167 with 1 strictly best, M$^2$WIS+S 150 with 3 strictly best, and HtWIS 135 with none strictly best. 
A similar pattern holds for `Avg'. ARCIS achieves the best or tied-best `Avg' on 201 instances, with 41 strictly best. The baselines achieve ReduMIS 171 best `Avg' with 7 strictly best, CHILS 165 with 4 strictly best, M$^2$WIS+S 130 with 4 strictly best, and HtWIS 135 with none strictly best. These results indicate that ARCIS attains superior solution quality in most instances in both peak and mean measures under a common time budget.

\begin{table}[!t]
\centering
\caption{Experimental results on the Walshaw benchmark.}
\label{tab:Walshaw}
\resizebox{0.9\linewidth}{!}{
\begin{tabular}{p{2cm}p{3cm}p{2cm}p{3cm}p{3cm}p{3cm}}
\toprule
Instance & ReduMIS & HtWIS & M$^2$WIS+S & CHILS & ARCIS \\
\midrule
144 & 26771(26759.5) & 25035 & 25154(25128.2) & 26805(26794.7) & \textbf{26849}(\textbf{26831.2}) \\
598a & 21765(21746.3) & 20228 & 20398(20369.8) & 21791(21781.5) & \textbf{21813}(\textbf{21806.3}) \\
auto & 83684(83621.5) & 78475 & 78566(78546.8) & 83942(83915.1) & \textbf{84081}(\textbf{84035}) \\
bcsstk29 & \textbf{1350} & 1323 & \textbf{1350} & \textbf{1350} & \textbf{1350}(1349.9) \\
bcsstk30 & \textbf{1783} & 1777 & \textbf{1783} & \textbf{1783} & \textbf{1783} \\
bcsstk31 & \textbf{3488} & 3477 & \textbf{3488} & \textbf{3488} & \textbf{3488} \\
bcsstk32 & \textbf{2967} & 2941 & \textbf{2967} & \textbf{2967}(2966.2) & \textbf{2967}(2966.2) \\
bcsstk33 & \textbf{512} & 504 & \textbf{512} & \textbf{512} & \textbf{512}(510.9) \\
brack2 & \textbf{21418}(21417.3) & 21385 & \textbf{21418}(21417.3) & \textbf{21418}(21415.6) & \textbf{21418}(\textbf{21417.6}) \\
crack & \textbf{4603} & 4601 & \textbf{4603} & \textbf{4603} & \textbf{4603} \\
cs4 & 9160(9156.6) & 8660 & 8943(8924) & \textbf{9167}(\textbf{9162.9}) & 9161(9156.7) \\
cti & \textbf{8088}(8076.1) & 7841 & \textbf{8088} & 8071 & \textbf{8088}(8084.6) \\
fe\_rotor & 21994(21975.9) & 19863 & 20992(20961.7) & 21999(21940.4) & \textbf{22025}(\textbf{22018}) \\
fe-body & \textbf{13750} & 13689 & 13734(13733.6) & \textbf{13750}(13749.4) & \textbf{13750} \\
fe-ocean & 71592(71204.7) & 67757 & \textbf{71716}(\textbf{71665.4}) & 71371(71269) & \textbf{71716}(71599.9) \\
fe-pwt & 9308(9305.7) & 8919 & 9294(9291.7) & \textbf{9309} & \textbf{9309}(9308.6) \\
fe-sphere & \textbf{5462} & 5328 & \textbf{5462} & \textbf{5462} & \textbf{5462} \\
finan512 & \textbf{28160} & \textbf{28160} & \textbf{28160}(28158.8) & \textbf{28160} & \textbf{28160} \\
t60k & 29682(29655) & 29475 & 29628(29605) & 29673 & \textbf{29714}(\textbf{29708.1}) \\
vibrobox & \textbf{1852}(1851) & 1789 & 1787(1775.1) & 1851(1850.2) & \textbf{1852}(\textbf{1851.6}) \\
wing & 25201(25183.6) & 23641 & 24342(24312.1) & \textbf{25251}(\textbf{25242.1}) & 25214(25208.3) \\
\bottomrule
\end{tabular}
}
\end{table}

\begin{table}[!t]
\centering
\caption{Experimental results on the SNAP benchmark.}
\label{tab:snap}
\resizebox{0.95\linewidth}{!}{
\begin{tabular}{llllll}
\toprule
Instance & ReduMIS & HtWIS & M$^2$WIS+S & CHILS & ARCIS \\
\midrule
amazon0302 & \textbf{93578} & 93560 & \textbf{93578} & 93577(93575.6) & \textbf{93578} \\
amazon0312 & \textbf{139158}(139157) & 139009 & 139011(139004) & 139148(139145.2) & \textbf{139158} \\
amazon0505 & 143009(143008.5) & 142869 & 142894(142881.7) & 143000(142998.4) & \textbf{143010}(\textbf{143009.3}) \\
amazon0601 & \textbf{136843} & 136729 & 136559(136551) & 136833(136831.1) & \textbf{136843} \\
cit-Patents & 2089996(2089960.6) & 2088088 & 2079556(2079497.3) & 2090093(2090066.4) & \textbf{2090566}(\textbf{2090546.4}) \\
web-BerkStan & \textbf{408480}(408475.6) & 408083 & 408476(408472.2) & 408154(408120) & 408479(\textbf{408476.5}) \\
web-Google & \textbf{529138} & 529129 & \textbf{529138} & 529137(529136.3) & \textbf{529138} \\
web-NotreDame & \textbf{251851}(\textbf{251849.7}) & 251783 & 251849(251844) & 251850(251847.1) & 251850(251849.4) \\
web-Stanford & \textbf{163390}(163388.7) & 163259 & 163378(163374.8) & 163297(163281) & \textbf{163390} \\
wiki-Talk & \textbf{2338222} & \textbf{2338222} & \textbf{2338222} & 2337883(2337853) & \textbf{2338222} \\
\bottomrule
\end{tabular}
}
\end{table}

\begin{table}[!t]
\centering
\caption{Experimental results on the DIMACS10 benchmark.}
\label{tab:dimacs10}
\resizebox{0.95\linewidth}{!}{
\begin{tabular}{llllll}
\toprule
Instance & ReduMIS & HtWIS & M$^2$WIS+S & CHILS & ARCIS \\
\midrule
333SP & 1153310(1143180) & 1084265 & 1130509(1130149.7) & 1167220(1167083) & \textbf{1168521}(\textbf{1168023.2}) \\
audikw\_1 & 50456(50420.5) & 47347 & 47422(47408.7) & 50566(50543.4) & \textbf{50638}(\textbf{50612.6}) \\
belgium\_osm & \textbf{724487} & 724485 & \textbf{724487} & 724470(724466.7) & \textbf{724487}(724486.9) \\
cage15 & 994401(988472.1) & 912364 & 975191(974941.8) & \textbf{1073158}(\textbf{1072714.2}) & 1052285(1051561.2) \\
caidaRouterLevel & \textbf{117150} & 117135 & \textbf{117150} & \textbf{117150}(117148.4) & \textbf{117150} \\
cnr-2000 & 230034(230023) & 229815 & 229784(229768.8) & 229961(229944.5) & \textbf{230035}(\textbf{230034.7}) \\
ecology1 & \textbf{500000}(499356.7) & \textbf{500000} & \textbf{500000} & \textbf{500000} & \textbf{500000} \\
G\_n\_pin\_pout & 31594(31554.6) & 30036 & 30157(30134.5) & 31549(31523.5) & \textbf{31613}(\textbf{31589.4}) \\
kron\_g500-logn16 & \textbf{34591} & \textbf{34591} & \textbf{34591} & \textbf{34591}(34589.9) & \textbf{34591} \\
kron\_g500-logn17 & \textbf{92249} & \textbf{92249} & \textbf{92249} & \textbf{92249}(92248.7) & \textbf{92249} \\
kron\_g500-logn18 & \textbf{189184} & \textbf{189184} & \textbf{189184} & 189178(189175.5) & \textbf{189184} \\
kron\_g500-logn19 & \textbf{387518} & \textbf{387518} & \textbf{387518} & 387490(387482.7) & \textbf{387518} \\
ldoor & \textbf{52783} & 52557 & \textbf{52783} & 52781(52779.6) & \textbf{52783} \\
preferentialAttachment & 42070(42061.7) & 41867 & 41498(41470.6) & 42090(42084.7) & \textbf{42099}(\textbf{42085.4}) \\
rgg\_n\_2\_17\_s0 & 25416(25414.5) & 24602 & \textbf{25423} & 25416(25413.7) & 25421(25420.5) \\
rgg\_n\_2\_19\_s0 & 93203(93182.8) & 88780 & 89170(89157.1) & 93292(93280.6) & \textbf{93332}(\textbf{93315.7}) \\
rgg\_n\_2\_20\_s0 & 178668(178626.8) & 169140 & 170549(170523.3) & 179099(179078) & \textbf{179239}(\textbf{179195.8}) \\
rgg\_n\_2\_21\_s0 & 342391(342212.4) & 322924 & 326886(324479.1) & 344423(344387) & \textbf{344891}(\textbf{344787.9}) \\
rgg\_n\_2\_22\_s0 & 652116(649895.9) & 617821 & 502267(501868.3) & 662633(662608.6) & \textbf{664523}(\textbf{664136.3}) \\
rgg\_n\_2\_23\_s0 & 1230719(1216649.3) & 1182909 & 961732(961554.1) & 1276064(1276009.3) & \textbf{1276902}(\textbf{1276637.1}) \\
smallworld & 24502(24475.6) & 22928 & 22841(22822.7) & 24576(24562.9) & \textbf{24594}(\textbf{24586.7}) \\
wave & 37015(36925.7) & 28859 & 36178(36104.3) & 36895(36671.6) & \textbf{37040}(\textbf{37027.7}) \\
\bottomrule
\end{tabular}
}
\end{table}

\begin{table}[!h]
\centering
\caption{Experimental results on the Network Repository benchmark.}
\label{tab:real-world}
\resizebox{\linewidth}{!}{
\begin{tabular}{llllll}
\toprule
Instance & ReduMIS & HtWIS & M$^2$WIS+S & CHILS & ARCIS \\
\midrule
amazon-2008 & 309793(309792.9) & 309593 & 309423(309414.9) & 309789(309785.5) & \textbf{309794}(\textbf{309793.7}) \\
eu-2005 & 452351(452334.6) & 451839 & 451654(451633.9) & 452236(452188.3) & \textbf{452352}(\textbf{452350.5}) \\
hugetrace-00000 & 2261222(2243507.9) & 2266682 & 2257458(2257261.6) & 2274707(2274469.7) & \textbf{2278367}(\textbf{2278264.5}) \\
hugetric-00000 & 2864574(2780543.8) & 2872432 & 2859950(2859242.3) & 2883808(2883479.6) & \textbf{2888983}(\textbf{2888789.9}) \\
ia-infect-hyper & \textbf{23} & 22 & \textbf{23} & \textbf{23} & \textbf{23} \\
in-2004 & \textbf{896762}(896760.6) & 896680 & \textbf{896762} & 896544(896507.9) & \textbf{896762}(896760.4) \\
inf-roadNet-CA & \textbf{957497}(\textbf{957460.8}) & 955393 & 956507(956474) & 957369(957356) & 957468(957459.5) \\
inf-roadNet-PA & \textbf{533357}(533315) & 532209 & 532944(532906.8) & 533291(533285.9) & 533337(\textbf{533335.1}) \\
inf-road-usa & 12427554(12427426.1) & 12418824 & 12423108(12423039.9) & 12425341(12425255) & \textbf{12427602}(\textbf{12427580.8}) \\
libimseti & 127286(126622.2) & 127264 & 127050(127034.8) & 127286(127280.4) & \textbf{127294}(\textbf{127293.8}) \\
rt-retweet-crawl & \textbf{1031662} & \textbf{1031662} & \textbf{1031662} & \textbf{1031662}(1031661.9) & \textbf{1031662} \\
sc-ldoor & \textbf{95449} & 95228 & \textbf{95449} & 95448(95446.9) & \textbf{95449} \\
sc-msdoor & \textbf{34305} & 34246 & \textbf{34305} & \textbf{34305} & \textbf{34305} \\
sc-nasasrb & 3632(3631) & 3441 & 3496(3492.5) & 3633(3631.5) & \textbf{3635}(\textbf{3633.8}) \\
sc-pkustk11 & \textbf{3893} & 3885 & \textbf{3893} & \textbf{3893}(3891.3) & \textbf{3893} \\
sc-pkustk13 & \textbf{5676}(5674.2) & 5544 & 5512(5509.9) & 5674(5672.1) & \textbf{5676}(\textbf{5675.6}) \\
sc-pwtk & 10218(10213) & 9998 & 10088(10078.7) & 10221(10219.7) & \textbf{10224}(\textbf{10221.2}) \\
sc-shipsec1 & 23573(23554.5) & 21920 & 22471(22450.4) & 23525(23505.3) & \textbf{23595}(\textbf{23586.2}) \\
sc-shipsec5 & \textbf{32373}(\textbf{32352.2}) & 31806 & 31305(31278.3) & 32267(32248.2) & 32357(32349.3) \\
soc-academia & \textbf{117069} & 117068 & \textbf{117069} & \textbf{117069}(117068.1) & \textbf{117069} \\
soc-BlogCatalog & \textbf{68032} & \textbf{68032} & \textbf{68032} & \textbf{68032}(68031.9) & \textbf{68032} \\
soc-buzznet & \textbf{70550} & \textbf{70550} & \textbf{70550} & 70547(70545.2) & \textbf{70550} \\
soc-delicious & \textbf{450810}(450809.7) & 450776 & \textbf{450810} & 450784(450768.8) & \textbf{450810} \\
soc-digg & \textbf{667565} & \textbf{667565} & \textbf{667565} & 667561(667559.2) & \textbf{667565} \\
socfb-A-anon & \textbf{2721935} & \textbf{2721935} & \textbf{2721935} & 2721933(2721928.5) & \textbf{2721935} \\
socfb-B-anon & \textbf{2634564} & \textbf{2634564} & \textbf{2634564} & 2634562(2634559.3) & \textbf{2634564} \\
socfb-Berkeley13 & \textbf{5689}(\textbf{5687.1}) & 5616 & 5630(5617.3) & 5687(5685.3) & \textbf{5689}(5686.6) \\
socfb-CMU & \textbf{1635} & 1613 & 1625(1621.6) & 1634 & \textbf{1635}(1634.2) \\
socfb-Duke14 & \textbf{2202} & 2176 & 2186(2180.1) & \textbf{2202}(2200.9) & \textbf{2202}(2201.4) \\
socfb-Indiana & \textbf{6416}(\textbf{6413.3}) & 6291 & 6311(6304) & 6415(6408.8) & \textbf{6416}(\textbf{6413.3}) \\
socfb-MIT & \textbf{1745} & 1732 & 1739(1734.9) & \textbf{1745} & \textbf{1745}(1744.3) \\
socfb-OR & 26844(26843.9) & 26808 & 26738(26731.4) & 26843(26841.9) & \textbf{26845}(\textbf{26844.8}) \\
socfb-UCLA & \textbf{5230}(\textbf{5229.4}) & 5161 & 5177(5171.5) & 5229(5224.9) & \textbf{5230}(5228.1) \\
soc-flickr & \textbf{360698} & 360696 & \textbf{360698} & \textbf{360698}(360697.7) & \textbf{360698}(360697.4) \\
soc-flickr-und & \textbf{1240618} & \textbf{1240618} & \textbf{1240618} & 1240617(1240615.4) & \textbf{1240618} \\
soc-FourSquare & \textbf{548906} & \textbf{548906} & \textbf{548906} & \textbf{548906}(548905.2) & \textbf{548906} \\
soc-google-plus & \textbf{105887}(105886.5) & 105867 & 105842(105840.9) & 105885(105884) & \textbf{105887}(\textbf{105886.9}) \\
soc-gowalla & \textbf{112369} & 112368 & \textbf{112369} & \textbf{112369}(112368.5) & \textbf{112369} \\
soc-lastfm & \textbf{1113117} & \textbf{1113117} & \textbf{1113117} & \textbf{1113117}(1113116.9) & \textbf{1113117} \\
soc-LiveJournal1 & \textbf{2631903} & 2631871 & \textbf{2631903} & 2631812(2631803.3) & \textbf{2631903} \\
soc-livejournal & \textbf{2164234} & 2164225 & \textbf{2164234} & 2164192(2164187.8) & \textbf{2164234} \\
soc-LiveMocha & \textbf{60676} & \textbf{60676} & \textbf{60676} & 60674(60672.9) & \textbf{60676} \\
soc-pokec & 787673(786977.4) & 789120 & 781381(781311.7) & 789235(789226.2) & \textbf{789460}(\textbf{789455.3}) \\
soc-twitter-follows-mun & \textbf{462517} & \textbf{462517} & \textbf{462517} & \textbf{462517}(462516.6) & \textbf{462517} \\
soc-twitter-follows & \textbf{402396} & \textbf{402396} & \textbf{402396} & \textbf{402396}(402395.6) & \textbf{402396} \\
soc-twitter-higgs & 249987(249985.2) & 249959 & 249580(249554.2) & 249637(249615.8) & \textbf{249991}(\textbf{249990.5}) \\
soc-wiki-Talk-dir & \textbf{2338222} & \textbf{2338222} & \textbf{2338222} & 2337865(2337837.3) & \textbf{2338222} \\
soc-youtube-snap & \textbf{857945} & \textbf{857945} & \textbf{857945} & \textbf{857945}(857944.2) & \textbf{857945} \\
soc-youtube & \textbf{349581} & \textbf{349581} & \textbf{349581} & \textbf{349581}(349580.6) & \textbf{349581} \\
tech-as-skitter & \textbf{1169594}(1169591.1) & 1169555 & 1169544(1169539.3) & 1168971(1168956.7) & \textbf{1169594}(\textbf{1169593.9}) \\
tech-RL-caida & \textbf{116321} & 116306 & \textbf{116321} & \textbf{116321}(116319.5) & \textbf{116321} \\
web-arabic-2005 & \textbf{49178} & 49177 & \textbf{49178} & \textbf{49178} & \textbf{49178} \\
web-it-2004 & \textbf{94831} & 94817 & \textbf{94831} & 94769(94763.5) & \textbf{94831} \\
web-sk-2005 & \textbf{63249} & 63239 & \textbf{63249} & \textbf{63249} & \textbf{63249} \\
web-webbase-2001 & \textbf{13411} & 13410 & \textbf{13411} & 13410 & \textbf{13411} \\
web-wikipedia2009 & \textbf{1216139}(1216137.1) & 1216077 & 1216077(1216067.6) & 1216126(1216121.4) & \textbf{1216139}(\textbf{1216137.5}) \\
\bottomrule
\end{tabular}
}
\end{table}

\subsection{Comparison of Convergence Speed}
To provide a fair comparison of convergence speed, we analyze running times only on instances where ARCIS and a given competitor achieve identical `Max' and `Avg' solution sizes. Following the protocol in~\cite{sunnumds}, we exclude any instance pair where both algorithms finished in under 0.1\,s to mitigate timing noise.

\begin{figure}[htbp]
	\centering
	\includegraphics[width=0.8\linewidth]{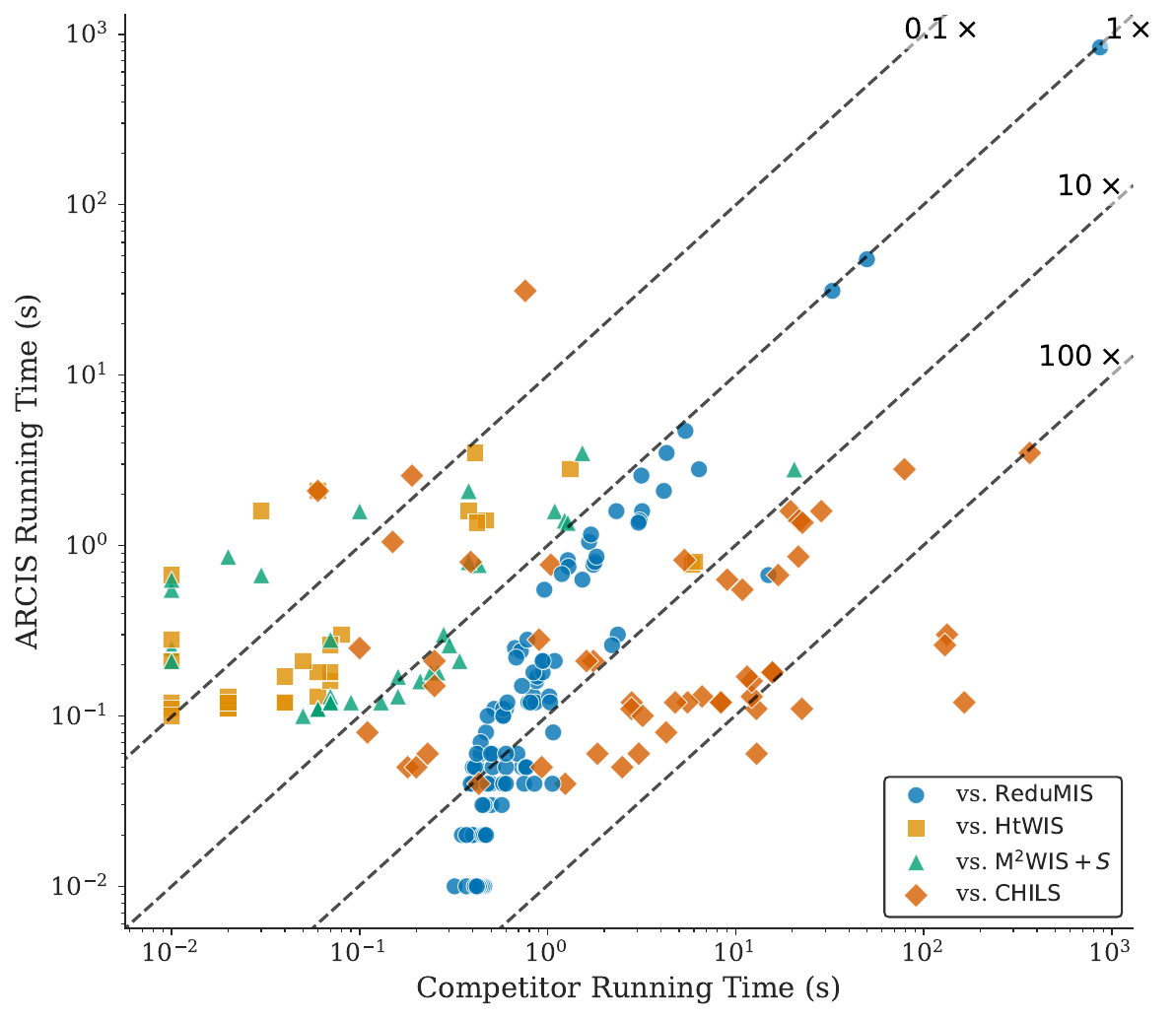}
	\caption{Comparison of run time of ARCIS and its corresponding competitor.}
	\label{fig:time}
\end{figure}

Figure~\ref{fig:time} presents this comparison as a scatter plot, with the running time of ARCIS on the vertical axis and the competitor's time on the horizontal axis. The plot includes four diagonal reference lines, labeled $0.1\times$, $1\times$, $10\times$, and $100\times$. Points falling \emph{below} the $1\times$ line (the main diagonal) represent instances where ARCIS is faster, while points \emph{above} this line indicate the competitor is faster. The other lines serve to illustrate the magnitude of the performance difference.

The plot shows that a clear majority of points fall below the $1\times$ line, indicating that ARCIS is faster than its competitors on these shared-solution instances. A smaller number of points lie above the diagonal, primarily in comparisons against HtWIS, which was faster on a subset of instances.

We further observe that the pairs on which HtWIS is faster coincide exactly with the subset where all algorithms produce identical Max and Avg across runs. 
This indicates that HtWIS tends to be faster when the solution landscape is relatively uniform across methods, whereas on pairs with differing solutions---a proxy for more challenging instances---ARCIS remains competitive or faster.

\subsection{Stability Analysis}

We assess stability across the four benchmarks (Walshaw, SNAP, DIMACS10, and Network Repository). HtWIS is excluded because in our environment it is executed only once per instance and does not rely on random seeds; hence seed-based variability is not applicable. The comparison includes ReduMIS, M$^2$WIS+S, CHILS, and ARCIS. For each instance and algorithm, we average independent-set sizes over ten seeds, then apply Min–Max normalization before aggregating across benchmarks.

\begin{figure}[htbp]
	\centering
	\includegraphics[width=\linewidth]{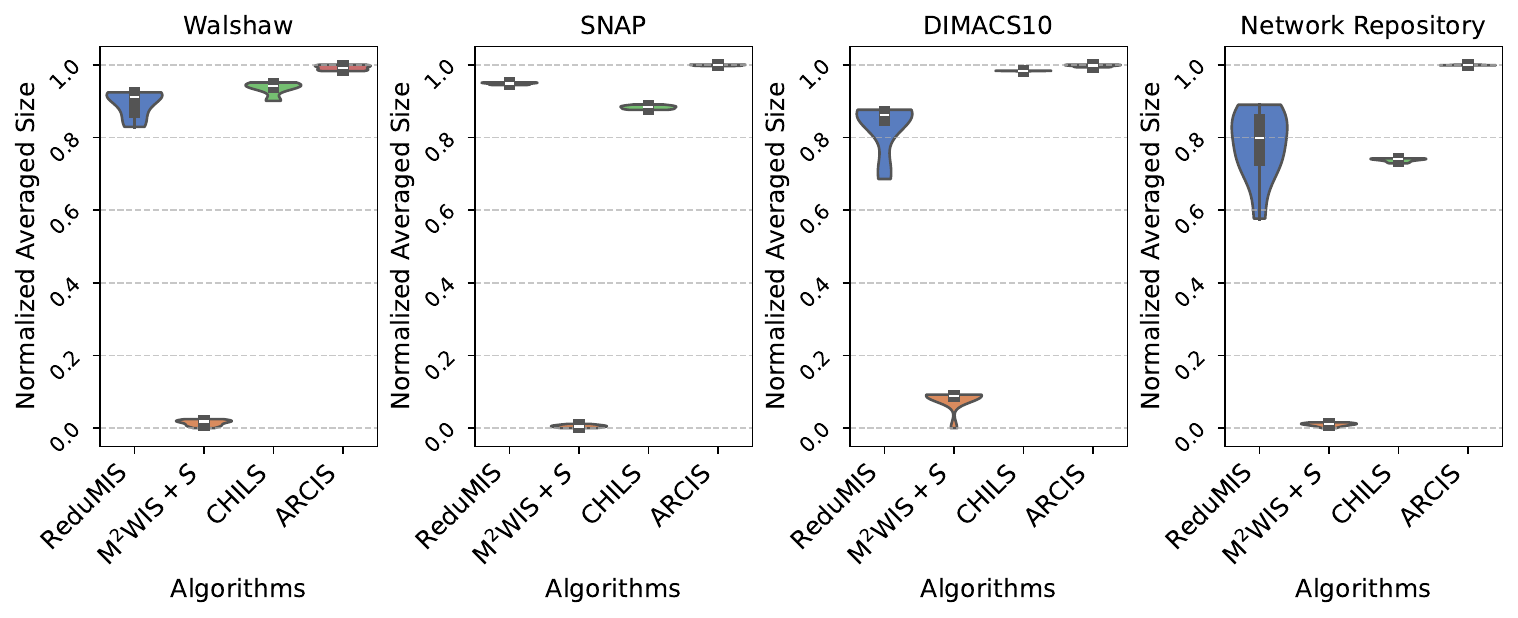}
	\caption{Stability analysis of algorithms.}
	\label{fig:stability}
\end{figure}

Figure~\ref{fig:stability} shows the distributions of these normalized averages for each algorithm. Tighter spreads indicate higher stability across seeds and instances, whereas wider spreads indicate greater variability. Higher central tendency corresponds to better average solution quality.
ARCIS exhibits the tightest dispersion with a higher central tendency across the four benchmarks, indicating both stable and strong performance.

\subsection{Statistical Analysis}
\label{sec:stats}

\begin{figure}[htbp]
    \centering
    \includegraphics[width=\linewidth]{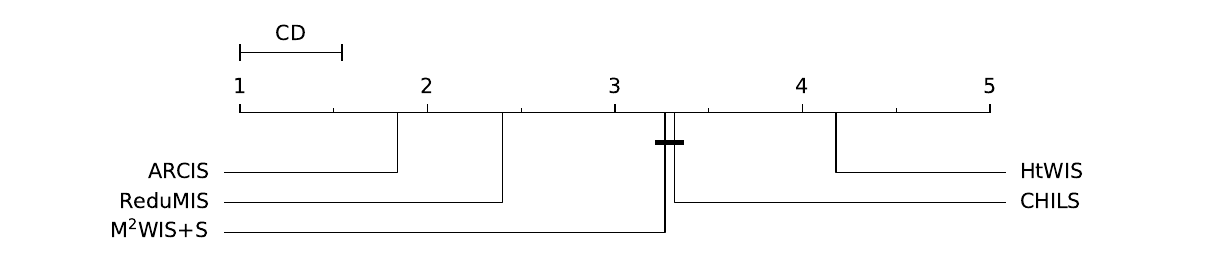}
    \caption{Critical-difference (CD) diagram using Friedman ranks with Nemenyi post-hoc at $\alpha=0.05$ (lower average rank is better). Bars connect methods that are not significantly different.}
    \label{fig:cd}
\end{figure}

To assess statistical significance, we apply the Friedman test on per-instance average ranks across all datasets, after excluding instances where all algorithms produced exactly the same solution. 
(For each instance, the score used for ranking is the mean independent-set size over 10 runs with seeds 1--10 for stochastic methods; HtWIS is deterministic and run once.)
The Friedman null hypothesis of equal performance~\cite{friedman1937use} is rejected at the $\alpha=0.05$ level, allowing a post-hoc Nemenyi test~\cite{demvsar2006statistical} to identify pairwise differences.
Figure~\ref{fig:cd} shows the resulting critical-difference (CD) diagram: methods are positioned by their average rank (lower is better), and horizontal bars connect methods that are \emph{not} significantly different under the Nemenyi test at $\alpha=0.05$.
In our plot, ARCIS attains the lowest average rank; the CD bars indicate which pairwise gaps are statistically reliable at the chosen level.

\subsection{Component Contribution via Ablation}

We compare ARCIS with two stripped variants that ablate key components. 
\textbf{ARCIS-I} removes both the adaptive restart policy and CGVF, while \textbf{ARCIS-II} keeps adaptive restarts but removes CGVF.
Since CGVF is applied only at restart boundaries, a variant that retains CGVF while removing \textsc{Evaluation} (which triggers restarts) is not applicable.

Table~\ref{tab:ablation_study} reports, for each method, the fraction of instances on which it yields a strictly larger independent set (`Better') or at least an equal size (`Greater or Equal') under both `Max' and `Avg'. ARCIS consistently obtains higher `Better' and `Greater or Equal' ratios than ARCIS-I and ARCIS-II, indicating that both the adaptive restart policy and CGVF contribute materially to overall performance.

\begin{table}[htb]
  \centering
    \caption{Comparison between ARCIS and two of its variants.}
    \resizebox{0.9\linewidth}{!}{
  \begin{tabular}{p{4cm}p{2cm}p{2cm}p{2cm}}
    \toprule
    Metric            & ARCIS & ARCIS-I & ARCIS-II \\
    \midrule
    Better (`Max')               & \textbf{31.53\%} & 17.56\% & 0.00\% \\
    Greater or Equal (`Max')     & \textbf{89.19\%} & 75.23\% & 42.34\% \\
    Better (`Avg')               & \textbf{45.95\%} & 29.28\% & 0.00\% \\
    Greater or Equal (`Avg')     & \textbf{80.63\%} & 60.81\% & 25.23\% \\
    \bottomrule
  \end{tabular}
  }
  \label{tab:ablation_study}
\end{table}

\section{Conclusion} \label{sec:conclusion}

We presented ARCIS, an efficient algorithm for the maximum independent set problem that combines an adaptive restart policy with Consensus-Guided Vertex Fixing. The restart policy reacts to observed progress and introduces diversification only when needed, and CGVF fixes vertices that persist across solutions within a round so that the next round searches only the remaining subgraph. The design is lightweight, reversible across rounds, and integrates cleanly with ARW-style improvement. On 222 graphs from four benchmark suites, ARCIS attains the best or tied-best solution quality on most instances under a common time budget, with competitive runtime and low variability across seeds. Ablation experiments further show that removing either component degrades performance, indicating that both the adaptive restart policy and CGVF contribute materially to the overall gains.

ARCIS is a heuristic algorithm and its behavior depends on a small number of controller parameters and on the checkpoint protocol. Although the consensus is conservative and reversible, occasional overcommitment can occur on difficult instances. Future work will investigate instance-adaptive schedules for the restart probability, confidence-weighted consensus updates, and parallel variants. We also plan to extend the algorithm to weighted and related independence problems and to explore hybrid designs in which learned signals guide restart probing or candidate selection while the core consensus mechanism and evaluation protocol remain unchanged.

\bibliographystyle{splncs04}
\bibliography{ref}

\end{document}